\documentclass[letterpaper, 10 pt, conference]{ieeeconf}  

\IEEEoverridecommandlockouts                              

\usepackage{cite}
\usepackage{mathptmx} 
\usepackage{amsmath} 
\usepackage{amssymb}  
\usepackage{adjustbox}
\usepackage{graphicx}
\usepackage[dvipsnames]{xcolor}
\usepackage{booktabs}
\usepackage{gensymb}
\usepackage{multirow}
\usepackage{hyperref}
\usepackage{arydshln}
\usepackage{svg}
\usepackage{adjustbox}
\usepackage{textcomp}
\usepackage{gensymb}
\definecolor{synthetic}{rgb}{0.0,0.0,0.8}
\definecolor{real}{rgb}{0.6,0.0,0.1}

\newcommand{\R}{\mathbb{R}}

\title{\LARGE \bf
DiffusionNOCS: Managing Symmetry and Uncertainty\\in Sim2Real Multi-Modal Category-level Pose Estimation
}

\author{Takuya Ikeda$^{*,1}$, Sergey Zakharov$^{*,2}$, Tianyi Ko$^{1}$, Muhammad Zubair Irshad$^{2}$, Robert Lee$^{1}$, \\ Katherine Liu$^{2}$, Rares Ambrus$^{2}$, Koichi Nishiwaki$^{1}$
\thanks{$^{1}$ All authors are with the Woven by Toyota. 3 Chome-2-1 Nihonbashimuromachi, Chuo City, Tokyo 103-0022, Japan, {\tt\small [firstname.lastname]@woven.toyota}}%
\thanks{$^{2}$ All authors are with the Toyota Research Institute, 4440 El Camino Real, Los Altos, CA 94022, USA, {\tt\small [firstname.lastname]@tri.global}}
\thanks{$^{*}$ Equal contribution. This work has been submitted to the IEEE for possible publication. Copyright may be transferred without notice, after which this version may no longer be accessible.}
}

\begin{document}

\maketitle
\thispagestyle{empty}
\pagestyle{empty}

\begin{abstract}
This paper addresses the challenging problem of category-level pose estimation. Current state-of-the-art methods for this task face challenges
when dealing with symmetric objects and when attempting to generalize to new environments solely through synthetic data training. 
In this work, we address these challenges by proposing a probabilistic model that relies on diffusion to estimate dense canonical maps crucial for recovering partial object shapes as well as establishing correspondences essential for pose estimation. 
Furthermore, we introduce critical components to enhance performance
by leveraging the strength of the diffusion models with multi-modal input representations.
We demonstrate the effectiveness of our method by testing it on a range of real datasets. Despite being trained solely on our generated synthetic data, our approach achieves state-of-the-art performance and unprecedented generalization qualities, outperforming baselines, even those specifically trained on the target domain. 
\end{abstract}

\section{INTRODUCTION}

In the fields of computer vision and robotics, object pose estimation is a fundamental problem.
Recently, approaches based on neural networks \cite{hodan2020epos, zakharov2019dpod, su2022zebrapose} have shown great progress for instance-level object pose estimation, quickly saturating the existing benchmarks~\cite{bopchallenge}. Consequently, a more challenging benchmark was proposed \cite{Wang2019-hy}, relaxing instance-level object pose estimation to category-level and evaluating the pose of unseen objects within a given category. While a significant number of methods have made progress in category-level pose estimation \cite{tian2020shape, Lin2021-gh, irshad2022centersnap}, many challenges such as handling ambiguity and generalization persist.
 
The challenges inherent in modelling the ambiguity of object pose are particularly apparent in symmetric objects, where a set of poses appear to be equally feasible given a partial observation. Most current methods regress a single pose when presented with a partially observed input, and therefore must craft methods to handle symmetry issues such as extensive data labeling and heuristic operations to avoid training instabilities that can stem from the underlying multi-modality of the pose. In some cases, ambiguity can be resolved given additional information beyond geometry. For example, adding an image-space observation of a laptop can help to disambiguate the keyboard from the screen \cite{You2022CPPFTR}. However, there are also common object instances for which visual texture cannot disambiguate the pose; consider the case of a uniformly colored cylindrical cup, where any rotation around the upright axis is a plausible pose candidate. Additionally, the existence and quality of different input modalities may change at inference time due to variations in sensor configuration and performance. We posit that utilizing a probabilistic method for pose estimation alongside 
multi-modal input representations can significantly improve the performance when ambiguities are present at deployment.

\begin{figure}[tbp]
\centering
\includegraphics[width=1\linewidth]{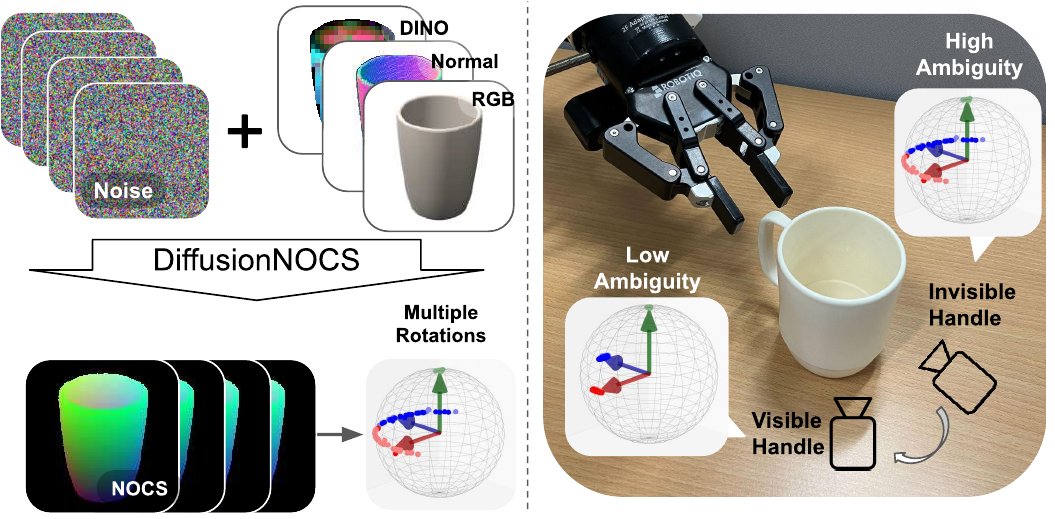}
\caption{~\textbf{Overview:} Our method can estimate multiple possible poses via diffusion models with a single observation. The diffusion models can be conditioned on any image, and naturally handle the ambiguity present from symmetry, where the predictions are certain when the mug handle is visible in the input image and are uncertain when it is invisible. The projected Normalized Object Coordinate Space(NOCS) map is denoised by the diffusion model and used as dense correspondences to estimate the poses.}
\label{fig:simple_overview}
\end{figure}

Generalization to the diversity within a class and across environments is another crucial challenge in building robust pose estimation. Collecting sufficient real-world training data for 3D pose estimation robust to visual texture and environment changes is a prohibitively consuming and costly process. Many existing works in the field of category-level pose estimation are trained on a specific set of real-world data and evaluated in a single environment~\cite{Wang2019-hy}.
Promising methods of mitigating the data scarcity issue include leveraging semantic features learned at scale for pose estimation \cite{goodwin2023you} and utilizing readily available labels from synthetic data \cite{You2022CPPFTR}. However, there is a notable lack of existing benchmarks suitable for evaluating the zero-shot performance of category-level pose estimation in novel, real-world settings.
 
In this work, we introduce a novel category-level dense pose estimation method based on diffusion models, with multi-modal input representations as shown in \autoref{fig:simple_overview}.
The multi-modal inputs and the probabilistic nature of our method allow it to naturally handle uncertainty, improving robustness and demonstrating state-of-the-art performance on various real datasets despite being trained solely on synthetic data.
The contributions of this paper are as follows:
\begin{enumerate}
    \item \label{contrib:pipeline} A novel pipeline for category-level object pose estimation
    via~\textbf{diffusion models with selectable multi-modal inputs}, including different combinations of RGB, surface normals and learned semantic features.
    Our approach's probabilistic nature enables it to seamlessly handle a range of symmetric objects without relying on hand-crafted heuristics.
    
    \item \label{contrib:eval} 
    A~\textbf{zero-shot generalization benchmark} demonstrating how existing state-of-the-art methods perform in challenging real-world environments featuring pose distribution, lighting, and occlusion changes.
    Despite being trained only on synthetic data, our proposed approach outperforms all baselines using synthetic and real data on average.
\end{enumerate} 

\section{Related Work}
\subsection{Object Pose Estimation}
Most existing pose estimation methods can be grouped into three categories: direct pose regression~\cite{kendall2015posenet, walch2017image, engelmann2020points}
(learning to output the pose directly from data), template-matching~\cite{hinterstoisserMultimodalTemplatesRealtime2011b,wohlhart2015learning, zakharov2017}
(comparing inputs with a predefined set of templates), and correspondence-based methods~\cite{zakharov2019dpod,hodan2020epos} (learning to estimate correspondences between input data and corresponding 3D models of interest) - each having their own benefits and drawbacks. Overall, the current state-of-the-art instance and category-level methods \cite{su2022zebrapose, Wang2019-hy, tian2020shape}
for object pose estimation are almost exclusively correspondence-based approaches.
Correspondence-based methods have clear benefits when handling occlusions and generalizing to different cameras. However, they typically require carefully curated datasets and hand-crafted operations to handle symmetrical objects to solve the ambiguity of 6D object pose. For example, Wang et al.~\cite{Wang2019-hy} designed loss functions for specific symmetry types. 
This is because the deterministic nature of their training cannot manage multiple feasible pose hypotheses.
Since real-world objects often exhibit symmetry, this poses a challenge when scaling to larger training datasets and applying these methods in the real world. 

\subsection{Handling Symmetry and Uncertainty}
While common pose estimation methods have shown great promise, they often lack the capability to provide information about uncertainty and are over-confident in their predictions~\cite{guo2017calibration}. In particular, they often rely on extensive prior knowledge of object geometry to address ambiguities coming from symmetry and occlusions. Such methods often require a dataset of symmetry-labeled objects~\cite{Wang2019-hy}, which is impractical in real-world applications. 
Another family of methods uses a probabilistic formulation that offers a means to capture uncertainty in the instance of interest directly.
Kendall et al.~\cite{kendall2017geometric} augmented PoseNet~\cite{kendall2015posenet} with uncertainty by sampling the posterior to approximate probabilistic inference. Bishop et al.~\cite{370fbeadb5584ba9ab2938431fc4f140} learned
to predict parameters of a Gaussian mixture distribution by using a neural network, which in turn can be used to infer the uncertainty based on the variance of the predicted distribution.
Gilitschenski et al.~\cite{gilitschenski2019deep} presented orientation uncertainty learning via Bingham distribution which is suited to represent the rotational ambiguity.
However, despite 
its effectiveness in representing axial symmetry, managing discrete symmetry is challenging
as the multi-modal distribution needs to be handled~\cite{deng2022deep}.
On the other hand, diffusion models offer a promising alternative, as these models can directly handle the multi-modality that is present in the training data \cite{ chi2023diffusion}.
Recently, Zhang et al.~\cite{zhang2023genpose} used a generative diffusion approach to estimate poses from partial point clouds, naturally handling all symmetry types. 
Our method also uses diffusion, but we instead estimate dense canonical maps from multi-modal image inputs that recover both pose and partial geometry,
which not only enables us to handle symmetry and other types of ambiguity arising from object occlusions and truncation but also provides robustness against the potential degradation of noisy input data.

\begin{figure*}[!ht]
\centering
\includegraphics[width=1.0\linewidth]{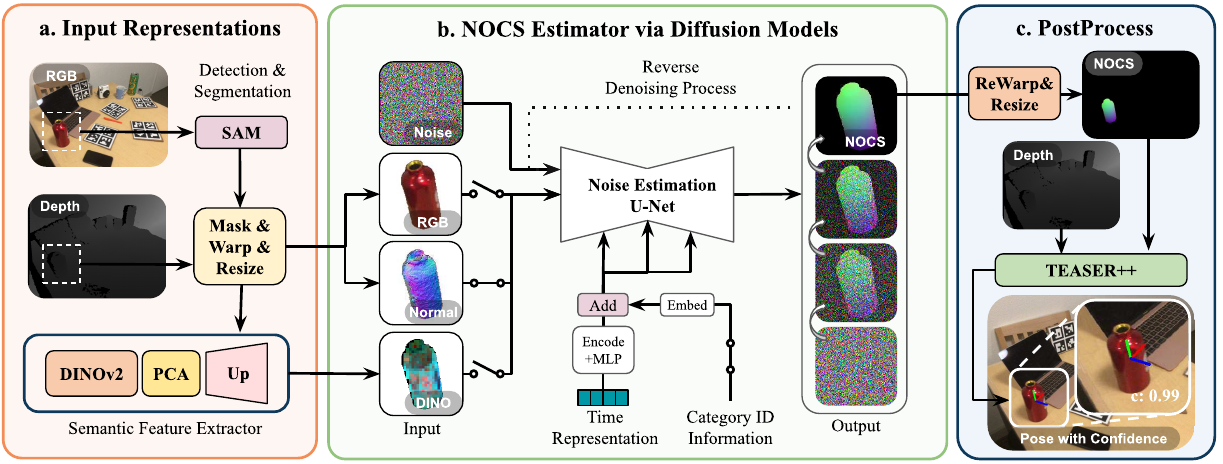}
\caption{\textbf{Pipeline Overview:} To prepare input representations 
from RGB, depth and given 2D bounding box, 
the operations of masking, warping, resizing are conducted sequentially. For the segmentation, SAM \cite{kirillov2023segment} with box prompts is used. To extract low-dimensional DINOv2 features with fixed shape, PCA and resizing are applied. Then, the diffusion model is conditioned by available inputs and estimates NOCS maps from noise images. Lastly, 6D pose and confidence are estimated via 
the point registration between denoised NOCS maps and partial point clouds from mask, depth, and given intrinsics~\cite{Yang20tro-teaser}.
}
\label{fig:overview}
\end{figure*}

\section{Method}  \label{method}
\textbf{Task Description.} We seek to estimate the 6D pose of unknown instances in known categories from a single image. At inference time, we assume a depth image, 2D bounding box and category ID encompassing detected objects are given, but can flexibly accommodate more inputs, such as RGB, if available.
To train the model, we use only synthetic data generated via rendering openly available 3D models, resulting in respective RGB, depth images, and category IDs. 

\textbf{Overview.} 
We propose a dense canonical map estimator via diffusion models that does not require special operations
to handle any type of symmetry.
We name our method DiffusionNOCS as the normalized object coordinate space (NOCS) \cite{Wang2019-hy} is leveraged as the dense canonical map, which our diffusion model is trained to predict. 
In this section, we first explain the network architecture and its input representations, then propose a synthetic data generation method and an entire pipeline for category-level object pose estimation.

\subsection{NOCS Preliminaries}
We utilize the NOCS as proposed in previous work \cite{Wang2019-hy}, where each object is canonicalized such that all points exist in some space ${x, y, z} \in [0, 1] $. The NOCS map is then the projection of the observed surface of the object into the camera frame. 
The resulting representation is an image-like quantity for which the channel dimension encodes the 3D coordinate of the surface point represented by the pixel. 
Given a predicted NOCS map, the pose of an object can be extracted via standard pose estimation techniques (i.e., PnP if only RGB is available and point cloud registration if depth information is available). 

\subsection{NOCS Estimator with Diffusion Models}
One of the key insights of this work is to leverage Denoising Diffusion Probabilistic Models (DDPMs) \cite{ho2020denoising} for NOCS map estimation, enabling the prediction of multiple possible maps via multiple noise samples.
Although DDPMs \cite{ho2020denoising} are commonly used for producing photo-realistic images from noise, we make two modifications for our purpose. 
First, we change the output from photo-realistic images to NOCS maps. Second, we condition the denoising process on images and category IDs. As such, we use the following loss for training:
\begin{equation}
    L = PixelwiseMSE(\epsilon^k, \epsilon_\theta(I, C, N^0 + \epsilon^k, k)),
\label{eq:loss}
\end{equation}
where $k$ represents the denoising iteration. $\epsilon^k$ is a random Gaussian noise map with proper variance for iteration $k$.  $N^0$ represents the raw NOCS maps, which are the ground truth from the dataset. $\epsilon_\theta$ is a noise estimation network with parameters $\theta$. $I$ and $C$ represent images and category IDs, respectively, for the conditioning of the noise estimator. 
For the training details, we refer reader to DDPMs \cite{ho2020denoising}. 
At the training phase, the discrete denoising scheduler \cite{ho2020denoising} with 1000 iterations is used as a noise scheduler.

The network overview is shown in the middle of \autoref{fig:overview}. Our main network is based on 2D U-Net \cite{ronneberger2015u} with ResNet blocks \cite{he2016deep}. Images to be conditioned are concatenated to the noise before being passed to the U-Net.
Additionally, two more components are utilized to manage the representation of time and category IDs. The first component is a sinusoidal positional encoding \cite{vaswani2017attention} and an MLP for time representation. The second is an embedding layer that serves as a 
lookup table for the category IDs. These outputs are then added and used as input for each layer of the U-Net model.

While the DDPM scheduler is utilized during training with 1000 iterations, repeating such a large number of iterations at inference time can be quite time-consuming. This is a significant concern for practical robotic applications where time efficiency is crucial.
As such, during inference we utilize the DPM-solver \cite{lu2022dpm}, which is a fast ODE-solver for diffusion models. This allows us to significantly decrease the iterations from 1000 to 10, improving the time efficiency to enable more effective use for real world robotics applications.

\subsection{Representations for Conditioning} 
\label{input_representations}
To estimate 6D poses for a large variety of objects with limited amount of real or synthetic data, the choice of input representation is critical. 
In our work, we choose to use four types of conditioning information: surface normals, RGB, DINOv2~\cite{oquab2023dinov2} semantic features, and category IDs.

\paragraph{Surface normals}
Depth information is essential for category-level pose estimation pipelines. Not only because it can help resolve scale-translation ambiguities, but also because it gives a good estimate of the partial object geometry. However, using unprocessed depth with learning-based methods poses certain challenges since it entangles scale, translation and camera intrinsic parameters, requiring large amounts of data to avoid overfitting.
In our work, we decouple estimation of these parameters and condition our NOCS estimator network on scale/translation invariant surface normals computed directly from input depth data. 

\paragraph{RGB images}
Geometric information alone can be highly ambiguous due to depth sensor's sensitivity to reflective objects, or when dealing with semantically rich but geometrically simple objects. 
For example, CPPF \cite{You2022CPPFTR} discusses the difficulty in distinguishing the lid of a laptop from the keyboard base from potentially noisy geometric information alone, such as partial point clouds. Therefore, the authors propose utilizing an additional network that takes RGB inputs. To be able to handle such cases, our method also uses RGB as a conditioning input.

\paragraph{DINOv2 features}
While RGB data can be highly useful, it is also a highly challenging modality. Even a single class of objects can contain a large variety of materials resulting in a diverse set of possible object appearances, while being very similar geometrically. Moreover, material appearance can further dramatically change under different lighting conditions. As a result, it requires large amounts of data to train methods that faithfully generalize to these changes, which can be very expensive.
Thankfully, powerful feature extractors trained on vast amounts of RGB data already exist and are readily available.
In our work, we leverage DINOv2 \cite{oquab2023dinov2}---a foundation model trained in a self-supervised manner, which proved to be useful for a variety of image-level visual tasks~\cite{goodwin2023you}.

\begin{figure}[tbp]
\centering
\includegraphics[width=1\linewidth]{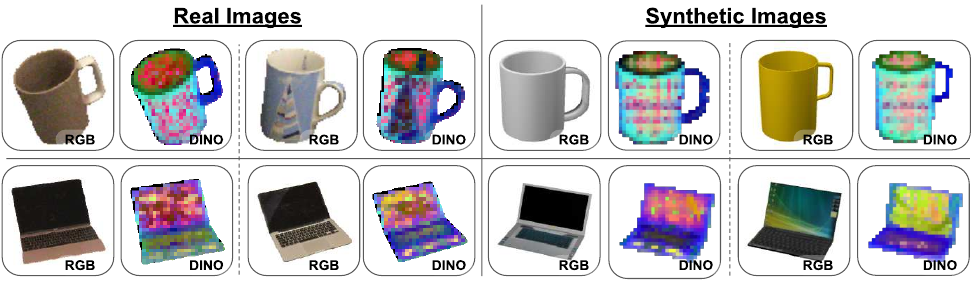}
\caption{~\textbf{Semantic Features}: 
DINOv2 features via 3-dimensional PCA with input RGB 
to understand the consistency across the different instances. }
\label{fig:dino_vis}
\end{figure}

To use DINOv2 features as a conditioning modality, we first preprocess them. The original feature dimension of the ViT-S-based DINOv2 is 384, which is much larger than all the previous modalities combined. This high-dimensional input not only makes inference and training speed slower, but also requires a larger network resulting in a larger memory footprint. To solve this problem, we follow Goodwin et al.~\cite{goodwin2023you} and utilize principal component analysis (PCA) to reduce the dimensionality from N to M (N$>$M).  The DINOv2 features via PCA with $N=384$, $M=3$ are visualized in \autoref{fig:dino_vis}, and show consistent coloring for different object instances from both real and synthetic images. 

\paragraph{Category IDs}
Category IDs are also included as an input modality.
They are readily available from
2D object detectors and serve as a valuable global semantic prior. 

\subsection{Selectable Input Representations}
\begin{figure}[htbp]
\centering
\includegraphics[width=0.9\linewidth]{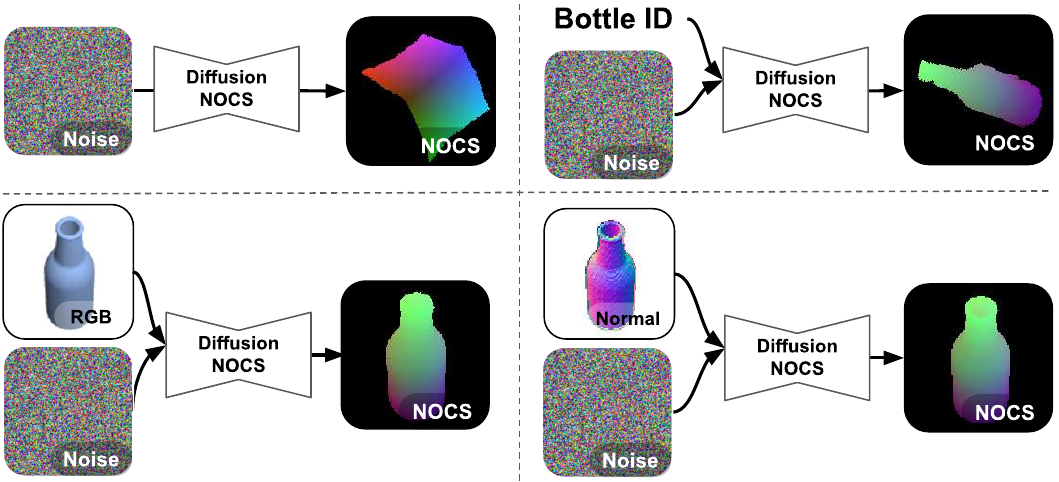}
\caption{ 
~\textbf{Estimated NOCS maps from Different Conditions:}
As a generative model, our method is capable of recovering plausible NOCS maps even when provided with only noise image input, i.e. the top left example shows a NOCS map highly resembling the "laptop" object.
Adding a category ID to the input limits the output distribution to the objects of a specified category (top right).
When RGB images or surface normals are provided, the silhouettes of the resulting NOCS maps are aligned with the inputs. Moreover, the output NOCS faithfully preserves geometric information provided by the surface normals as can be seen from the bottom right example where a hollow bottle is reconstructed. In contrast, the filled bottle is recovered when RGB information alone is provided (bottom left).
}
\label{fig:selectable_inputs}
\end{figure}

For the learning-based category-level pose estimation, depth only or RGB-depth is a common input representation.
To make our method more flexible, we propose an approach that can freely take the available inputs at inference time with a single network without re-training. 
To do this, we drop each condition at a certain rate during training as follows:  
\begin{equation}
    \Tilde{\epsilon}_\theta = 
    \epsilon_\theta(I_{normal} \lor \emptyset, I_{RGB} \lor \emptyset, I_{dino} \lor \emptyset, C \lor \emptyset, N^0 + \epsilon^k, k)
\label{eq:empty}
\end{equation}
$I$ in Eq. \ref{eq:loss} is separated to $I_{normal}$, $I_{RGB}$, and $I_{dino}$. When a condition is dropped, it is replaced with a \textit{null} value $\emptyset$, which in practice means a tensor of the same dimension as the original condition, but filled with zeros.
For example, dropping an image conditioning variable means that all pixels are filled by zeroes, and we let the zeroth ID represent an unknown category. Even if all conditioning modalities are dropped, the model attempts to generate NOCS maps from only noise, i.e., it performs unconditional diffusion. On the other hand, if an RGB image is available, the model generates NOCS by conditioning on the RGB image as shown in \autoref{fig:selectable_inputs}. 
At inference time, available inputs can be used without re-training. 
We observe this conditioning method is both powerful and flexible, enabling new useful representations to be incorporated as they are developed.

\subsection{Synthetic Data Generation} \label{data_gen}
\begin{figure}[htbp]
\centering
\includegraphics[width=0.9\linewidth]{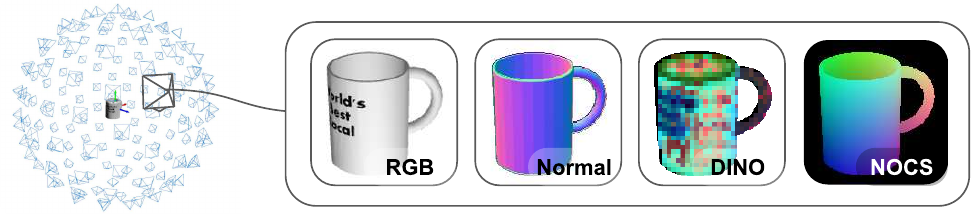}
\caption{~\textbf{Synthetic Data for DiffusionNOCS:} A visualization of camera poses and generated synthetic images used to train DiffusionNOCS.}
\label{fig:dat_gen}
\end{figure}

To make DiffusionNOCS generalize across various real environments using just a single dataset,
we introduce a domain-agnostic synthetic data generation method.
We assume 3D models with aligned coordination system are given.
RGB and depth images from 162 camera poses using the 3D models are rendered on white background as shown in \autoref{fig:dat_gen}. 
The camera poses are created via icosahedron sampling with 2 subdivision levels.
Then, we create normal and groundtruth NOCS maps given rendered depth as well as the known poses and sizes of objects. 
We estimate the DINOv2 features for all RGB images in the training dataset and apply principle component analysis (PCA) to reduce the dimension. 
As the resolution of DINOv2 features becomes lower than input RGB, we resize with nearest neighbor interpolation to ensure consistent dimensions.
To aid in generalization, we augment the data in three ways. 
The first is to apply in-plane rotation for the images at the training time instead of making camera poses more various at rendering time. 
The second is to apply lighting augmentation based on Phong reflection model \cite{phong1998illumination} with the normal maps at training time. 
Because of this, the lighting variation in the rendered RGB images can be less since it can be added after the rendering.
The third is conducting Cutout~\cite{devries2017cutout} by random shapes that are rectangle, circle, triangle, and ellipse to improve the robustness against occlusion.
At inference time, we also apply a warping operation to move the object to the center of image to ensure consistency with the training images.
We show how well the generalization of DiffusionNOCS trained by the synthetic data works across various real environments in \autoref{sota_comp}.

\subsection{Pipeline for Category-Level Pose Estimation} \label{pipeline}
The overview of entire inference pipeline for the 6D pose estimation is described in Fig. \ref{fig:overview}. 
For the condition of DiffusionNOCS, we aim to prepare fixed square shape normal maps, RGB images, and DINOv2 features with white background from given RGB, depth images and intrinsic parameters with 2D bounding box from detection. 
To create the normal maps and RGB images with fixed square shape, the cropping, warping and resizing operations are conducted sequentially on given RGB and depth images.
For the cropping and warpping, detected square bounding boxes and their center are used. To generate segmentations, a pretrained SAM~\cite{kirillov2023segment} with 2D bounding box prompts is used.
For the creation of low-dimentional DINOv2 features, the processed RGB image are fed to DINOv2 network, then PCA and resizing to the same size of the RGB image are applied. 
After the NOCS map estimation via DiffusionNOCS, the square NOCS maps are reverted to the original image shape by re-warping and resizing, then a point cloud registration is conducted between points from masked depth image with intrinsic and points from NOCS maps to estimate 6D pose and scale. TEASER++, a fast point cloud registration algorithm \cite{Yang20tro-teaser}, is utilized for the registration. After the registration, we treat the inlier rate from TEASER++ as the confidence of the estimated poses. This confidence is an important value for the next section.

\subsection{Handling Uncertainty for Robust Estimation} \label{boost}
\begin{figure}[t]
\centering
\includegraphics[width=1.0\linewidth]{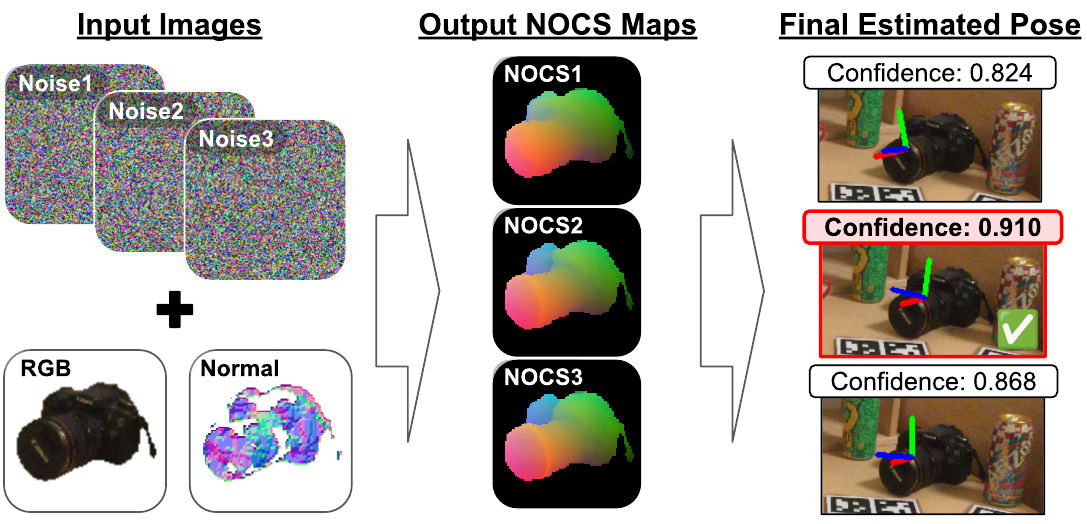}
\caption{\textbf{Final Pose Selection:}
By different noise images, multiple NOCS maps can be estimated.
TEASER++ predicts poses and their confidence by the NOCS maps. The final pose is selected based on the confidence value.}
\label{fig:selection}
\end{figure}

We propose an effective approach to boost the performance of 6D pose estimation at inference time via multiple noise sampling.
DiffusionNOCS can estimate multiple possible NOCS maps from the different input noises, even if with the same conditioning.
In other words, possible multiple poses can be estimated via multiple noise inputs as shown in \autoref{fig:selection}.
At inference time, we can input multiple noises as a batch, then select one estimated pose with the highest confidence as a final result.
In this work, we take the inliers rate from the TEASER++ rotation estimate as the confidence value.
This approach significantly improves the robustness of pose estimation, as shown in \autoref{tab:ablation1}.
Moreover, multiple estimated poses can be leveraged by various downstream tasks, such as grasping and placing. For example, if multiple high-confidence poses exist, the best grasping or placing pose can be selected by considering collision or kinematics constraints. 
As another example, to grasp the mug handle properly, the shape of rotation distribution can be used to measure the pose ambiguity as shown in \autoref{fig:simple_overview} and \autoref{fig:symmetry}.

\section{Experiments} \label{experiments}

\subsection{Experimental Setup}

\textbf{Benchmarks.} We evaluate our method on two benchmarks comprising a total of four real-world datasets, namely NOCS Real 275\cite{Wang2019-hy}, TYO-L~\cite{hodan2018bop}, YCB-V~\cite{xiang2018posecnn}, and HOPE~\cite{lin2021fusion}. 

The \textbf{NOCS Real Benchmark}~\cite{Wang2019-hy} is a de facto standard benchmark for category-level pose estimation. It contains objects of 6 target categories (bottle, can, bowl, laptop, camera, mug) placed in 6 different environments spanning a total of 2.75K test images. 
For a fair comparison to previous works \cite{zhang2023genpose, tian2020shape}, we use 2D bounding boxes and category IDs from precomputed MaskRCNN results as input during inference. The fine masks are estimated by pretrained SAM~\cite{kirillov2023segment} with the 2D bounding box prompts.

\begin{figure}[b]
\centering
\includegraphics[width=1.0\linewidth]{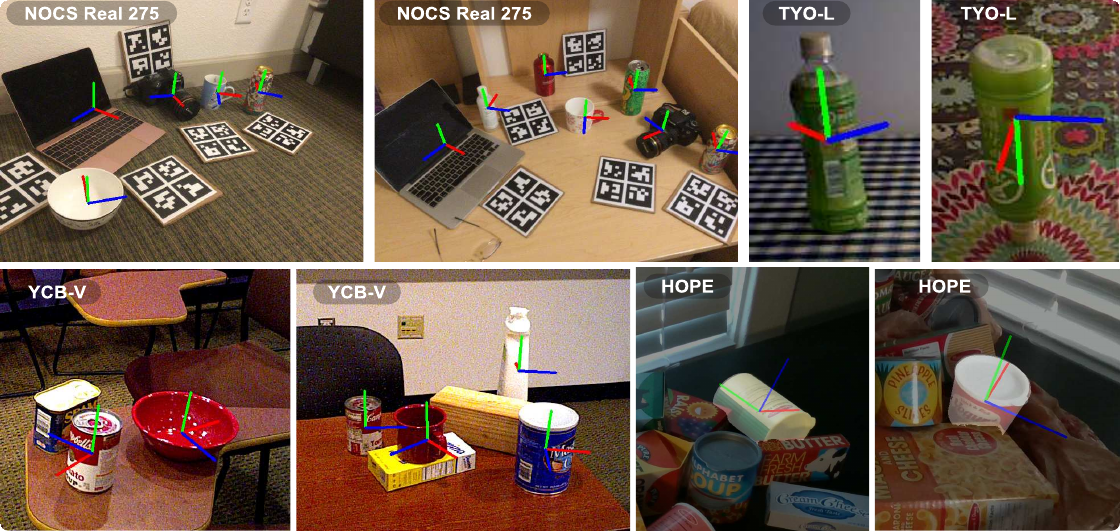}
\caption{\textbf{Visualization of estimated poses on NOCS Real 275, TYO-L, YCB-V, and HOPE datasets:} Despite training only on background-free synthetic data, we ~\textbf{show remarkable zero-shot generalization} on datasets featuring diverse pose, lighting, and occlusion distributions. HOPE dataset masks are colored as translucent white to highlight the occluded parts.}

\label{fig:pose_visualization}
\end{figure}

To demonstrate how existing state-of-the-art (SOTA) methods perform in various challenging real-world environments, we introduce a zero-shot \textbf{Generalization Benchmark} consisting of three datasets commonly used for instance-level pose estimation, TYO-L~\cite{hodan2018bop}, YCB-V~\cite{xiang2018posecnn}, and HOPE~\cite{lin2021fusion}. Each dataset poses different challenges to the baselines. TYO-L dataset features 21 single-object scenes with five different lighting conditions and four different backgrounds. HOPE dataset contains 10 highly occluded multi-object scenes with a very diverse pose distribution. The YCB-V dataset contains 12 multi-object scenes with light occlusions and is most similar to NOCS in terms of pose distribution. 
Figure~\ref{fig:pose_visualization} shows sample images from the used datasets.
To ensure fair evaluation, we standardize object models to be consistently canonicalized with the NOCS dataset and only use the object categories present in NOCS, resulting in 4 categories for TYO-L and YCB-V (bottle, bowl, mug, can), and 2 categories for HOPE (bottle, can). We utilize the GT masks for all baselines and ours during inference.

\textbf{Data Generation.} For training, we exclusively use synthetic data rendered using the Pyrender~\cite{pyrender} library. To ensure a non-biased class distribution, we render an equal number (20) of 3D models per target category from the ShapeNet dataset~\cite{shapenet2015}, which is the subset of 3D models used in NOCS CAMERA's training dataset. As described in \ref{data_gen},
cameras are sampled using the icosahedron sampling resulting in 162 images per instance. 
The all images have a size of 160 x 160 pixels. 

\textbf{Metrics.} Following original NOCS~\cite{Wang2019-hy} protocol, we use the mean Average Precision (mAP) in n$^{\circ}$ and m $cm$ to evaluate the accuracy of estimated poses. For axial-symmetric objects, the rotation error around the axis of symmetry is ignored.
In the generalization benchmark, we additionally adjust the metric for the ``can" category, the only object in the dataset that exhibits both axial and reflective symmetry, to be invariant to the direction of the axis of symmetry
since our benchmark contains objects in arbitrary poses, whereas NOCS Real 275 contains only objects standing upright. 
\begin{equation}
    \theta_{error} = \arccos\left(\dfrac {|Y_{GT} \cdot Y_{pred}|} {\left\| Y_{GT}\right\| _{2}\left\| Y_{pred}\right\| _{2}} \right), \quad Y_{GT}, Y_{pred} \in \R^{3}
\end{equation}
$Y_{GT}$ and $Y_{pred}$ represent the unit vector of the Y-axis which is aligned with the symmetry axis.

\begin{table}[b]
  \centering
  \caption{\textbf{Evaluation Results on NOCS REAL275.} 
    \textbf{Data} means the type of training data. \textcolor{synthetic}{\textbf{S}} denotes \textcolor{synthetic}{synthetic} ShapeNet objects only, while \textbf{S(B)} denotes ShapeNet models rendered with real backgrounds. \textcolor{real}{\textbf{R}} denotes \textcolor{real}{real} images in NOCS REAL275. $N$ and $DINO$ represent Normal and DINOv2 features respectively.}
    \resizebox{1.0\linewidth}{!}{
    \begin{tabular}{c|cc|ccc}
    \toprule
          & \textbf{Input} & \textbf{Data}  & \textit{$5 \degree 5$cm $\uparrow$} & \textit{$10 \degree 5$cm $\uparrow$} & \textit{$15 \degree 5$cm $\uparrow$} \\
    \hline
    NOCS~\cite{Wang2019-hy} & RGB-D & \textcolor{real}{S(B)+R}   & 9.8   & 24.1  & 34.9 \\
    ShapePrior~\cite{tian2020shape} & RGB-D & \textcolor{real}{S(B)+R}   & 21.4  & 43.6  & 66.7 \\
    CenterSnap~\cite{irshad2022centersnap} & RGB-D     & \textcolor{real}{S(B)+R}   & 27.2  & 58.8  & - \\
    DualPoseNet \cite{Lin2021-gh} & RGB-D & \textcolor{real}{S+R}   & 35.9  & 66.8  & 76.3 \\
    ShaPO~\cite{irshad2022shapo} & RGB-D & \textcolor{real}{S(B)+R}   & 48.8  & 66.8  & - \\
    GenPose \cite{zhang2023genpose} & D     & \textcolor{real}{S(B)+R}   & \textcolor{real}{\textbf{50.6}}  & \textcolor{real}{\textbf{84.0}}    & \textcolor{real}{\textbf{89.7}} \\
    \hline
    Chen et al. \cite{Chen2020CategoryLO} & RGB-D & \textcolor{synthetic}{S}     & 0.7   & 3.6   & 9.1 \\
    Gao et al. \cite{Gao20206DOP} & D     & \textcolor{synthetic}{S}     & 7.8   & 17.1  & 26.5 \\
    CPPF \cite{You2022CPPFTR} & D     & \textcolor{synthetic}{S}     & 16.9  & 44.9  & 50.8 \\
    ShapePrior \cite{tian2020shape} & RGB-D & \textcolor{synthetic}{S}     & 12.0  & 37.9  & 52.8 \\
    \hdashline
    Ours (N) & D     & \textcolor{synthetic}{S}     & 28.7  & 56.3  & 65.5 \\
    Ours (RGB-N) & RGB-D & \textcolor{synthetic}{S}     & 33.5  & 62.8  & 72.2 \\
    Ours (DINO-N) & RGB-D & \textcolor{synthetic}{S}     & 34.5  & \textcolor{synthetic}{\textbf{66.7}}  & 77.0 \\
    Ours (RGB-DINO-N) & RGB-D & \textcolor{synthetic}{S}     & \textcolor{synthetic}{\textbf{35.0}}  & 66.6  & \textcolor{synthetic}{\textbf{77.1}} \\
    \bottomrule
    \end{tabular}
    }
  \label{tab:nocs_eval}%
\end{table}%

\textbf{Implementation Details.}
At training time, we randomly drop the input representation at a 25\% rate via the procedure described by  Eq.~\ref{eq:empty}, and the time steps of DDPM are set to 1000. 
We apply PCA to DINOv2 features, reducing their dimension from 382 to 6 for both training and inference. At inference time, we use 6 different noises to generate 6 possible NOCS maps. Then, a final pose is selected based on the highest confidence.
Our choices of PCA dimension and the number of noises are informed by the ablation shown in \autoref{tab:ablation1}.
We use 10-time steps of a DPM-solver \cite{lu2022dpm}, which takes 0.855$s$ on the Tesla V100 GPU (corresponding to the 5th row in \autoref{tab:ablation1}). As noted in \autoref{tab:ablation1}, other faster configurations may be chosen as a trade-off with accuracy.

\subsection{Comparisons to State of the Art} \label{sota_comp}

\setlength{\tabcolsep}{5pt}
\begin{table*}[tbp]
  \centering
  \caption{\textbf{Evaluation Results on Generalization Benchmark.} The results show the zero-shot pose estimation performance on TYO-L, YCB-V, and HOPE real-world datasets. Our method outperforms all methods using \textcolor{synthetic}{synthetic data} and \textcolor{real}{real data} on average. 
  }
    \resizebox{1.0\textwidth}{!}{
    \begin{tabular}{c|c|ccc|ccc|ccc||ccc}
    \toprule
          & \multirow{2}[1]{*}{\textbf{Data}} & & \textbf{TYO-L} & & & \textbf{YCB-V} & & & \textbf{HOPE} & & & \textbf{Mean $\uparrow$ $\pm$ Std$\downarrow$} \\
\cline{3-14}          &       & \textit{$5 \degree 5$cm$\uparrow$} & \textit{$10 \degree 5$cm$\uparrow$} & \textit{$15 \degree 5$cm$\uparrow$} & \textit{$5 \degree 5$cm$\uparrow$} & \textit{$10 \degree 5$cm$\uparrow$} & \textit{$15 \degree 5$cm$\uparrow$} & \textit{$5 \degree 5$cm$\uparrow$} & \textit{$10 \degree 5$cm$\uparrow$} & \textit{$15 \degree 5$cm $\uparrow$} & \textit{$5 \degree 5$cm} & \textit{$10 \degree 5$cm} & \textit{$15 \degree 5$cm } \\
    \hline
    ShapePrior~\cite{tian2020shape} & \textcolor{real}{S+R}   & 14.9  & 27.5  & \textcolor{real}{\textbf{32.1}} & 27.4 & 54.3  &  66.9 & 14.1  & \textcolor{real}{\textbf{41.0}}  & \textcolor{real}{\textbf{57.0}}  & 18.8$\pm$\textcolor{real}{\textbf{6.1}} & \textcolor{real}{\textbf{40.9}}$\pm$\textcolor{real}{\textbf{10.9}} & \textcolor{real}{\textbf{52.0}}$\pm$\textcolor{real}{\textbf{14.6}} \\
    DualPoseNet \cite{Lin2021-gh} & \textcolor{real}{S+R} & \textcolor{real}{\textbf{22.5}}     & \textcolor{real}{\textbf{28.8}}     & 32.2  & 36.6     & \textcolor{real}{\textbf{78.8}}     & \textcolor{real}{\textbf{87.5}}    &   3.9    &   9.6    &   14.1    &   21.0$\pm$13.4    &   39.1$\pm$29.2   &  44.6$\pm$31.2 \\
    GenPose \cite{zhang2023genpose} & \textcolor{real}{S+R} & 20.1  & 23.7  & 24.7 & \textcolor{real}{\textbf{48.1}}  & 68.8  & 73.8 & \textcolor{real}{\textbf{18.8}}  & 23.2  & 25.2  & \textcolor{real}{\textbf{29.0}}$\pm$13.5 & 38.6$\pm$21.4 & 41.2$\pm$23.0 \\
    \hline
    CPPF \cite{You2022CPPFTR} & \textcolor{synthetic}{S}     & 13.2     & 18.8     & 20.8     & 8.8     & 35.2     & 46.9     &   3.2    &   15.3    &   20.5    &   8.4$\pm$ 4.1   &  23.1 $\pm$ 8.7 &  29.4 $\pm$ 12.4\\
    ShapePrior \cite{tian2020shape} & \textcolor{synthetic}{S}      & 17.3   & 26.5   & 33.4  & 11.0  & 40.3  & 49.1 &  15.0  & 26.8  & 32.7  & 14.4$\pm$\textcolor{synthetic}{\textbf{2.6}}  & 31.2$\pm$6.4 & 38.4$\pm$7.6 \\
    \hdashline
    Ours (N) & \textcolor{synthetic}{S}    & 33.8  & 43.9  & 49.4 & 20.9  & 46.2  & 53.0 & 17.9  & 37.3  & 45.7  & 24.2$\pm$6.9 & 42.5$\pm$\textcolor{synthetic}{\textbf{3.8}} & 49.4$\pm$\textcolor{synthetic}{\textbf{3.0}} \\
    Ours (RGB-N) & \textcolor{synthetic}{S}      & 40.3  & 51.0  & 56.6 & 20.0  & 47.3  & 51.0  & 19.1  & 40.0  & 46.8  & 26.5$\pm$9.8 & 46.1$\pm$4.6 & 51.5$\pm$4.0 \\
    Ours (DINO-N) & \textcolor{synthetic}{S}      & 43.5  & \textcolor{synthetic}{\textbf{60.2}}  & \textcolor{synthetic}{\textbf{66.0}}  & 23.4  & \textcolor{synthetic}{\textbf{54.8}}  & \textcolor{synthetic}{\textbf{58.2}}  & 20.1  & \textcolor{synthetic}{\textbf{40.2}}  & \textcolor{synthetic}{\textbf{47.1}}  & 29.0$\pm$10.3 & \textcolor{synthetic}{\textbf{51.7}}$\pm$8.4 & \textcolor{synthetic}{\textbf{57.1}}$\pm$7.8 \\
    Ours (RGB-DINO-N) & \textcolor{synthetic}{S}     & \textcolor{synthetic}{\textbf{43.7}}  & 59.0  & 64.7   & \textcolor{synthetic}{\textbf{23.6}}  & 53.7    & 56.6  & \textcolor{synthetic}{\textbf{20.4}}  & 38.5  & 46.5  & \textcolor{synthetic}{\textbf{29.2}}$\pm$10.3 & 50.4$\pm$8.7 & 55.9 $\pm$7.4 \\
    \bottomrule
    \end{tabular}%
    }
  \label{tab:gen_eval}%
\end{table*}%

\textbf{NOCS Real 275 Benchmark.} 
\autoref{tab:nocs_eval} reports a comparison between our method and SOTA approaches.
Our method shows significantly better results when compared to all other methods trained on synthetic data, achieving mAPs of 35.0 ($5 \degree 5$cm), 66.7 ($10 \degree 5$cm), and 77.1 ($15 \degree 5$cm), whereas the former best methods score 16.9 ($5 \degree 5$cm), 44.9 ($10 \degree 5$cm), and 52.8 ($15 \degree 5$cm) respectively.
Even though only synthetic data is used for our training, our method also demonstrates strong results when compared to the methods trained on a combination of both real and synthetic datasets outperforming NOCS~\cite{Wang2019-hy}, ShapePrior~\cite{tian2020shape}, and CenterSnap~\cite{irshad2022centersnap}.

We also show the importance of multi-modal input representations. The combination of surface normals and DINO features shows over 10 percent improvement compared with only normals on both 10$\degree$5cm and 15$\degree$5cm metrics. Also, \autoref{tab:nocs_eval} shows the DINO features are more effective than RGB.
For clarity, a single network was used to generate all score sets (N, RGB-N, DINO-N, RGB-DINO-N) without re-training since our method supports selectable inputs.

\textbf{Generalization Benchmark.} \label{gen_bench}   
To analyze zero-shot generalization capabilities of existing SOTA baselines, we computed their mAP scores on our newly introduced generalization benchmark as shown in \autoref{tab:gen_eval}. For all baselines, we used respective official implementations and provided pre-trained weights.
Similarly to the NOCS benchmark, we significantly outperform the SOTA synthetic baseline ShapePrior~\cite{tian2020shape} scoring average mAPs of 29.2 ($5 \degree 5$cm), 51.7 ($10 \degree 5$cm), and 57.1 ($15 \degree 5$cm) against ShapePrior's 14.4 ($5 \degree 5$cm), 31.2 ($10 \degree 5$cm), and 38.4 ($15 \degree 5$cm) in rightmost block in \autoref{tab:gen_eval}. 
Moreover, we outperform all real methods when averaging over all three datasets, 
demonstrating our robustness to occlusions, changing lighting conditions, and a wide variety of object poses. \autoref{tab:gen_eval} also shows real methods dominate the YCB-V scores.
We explain this by this dataset's similarity with NOCS: it only contains objects standing upright placed in well-lit scenes placed close to the camera. 
Overall, as the mean and standard deviation scores demonstrate, our method achieves the best overall performance and stability even when compared to the methods trained on synthetic and real data.

\subsection{Analysis on Symmetric Objects}
\begin{figure}[t]
\centering
\includegraphics[trim={0.0cm 0.0cm 0cm 0cm},clip, width=0.95\linewidth]{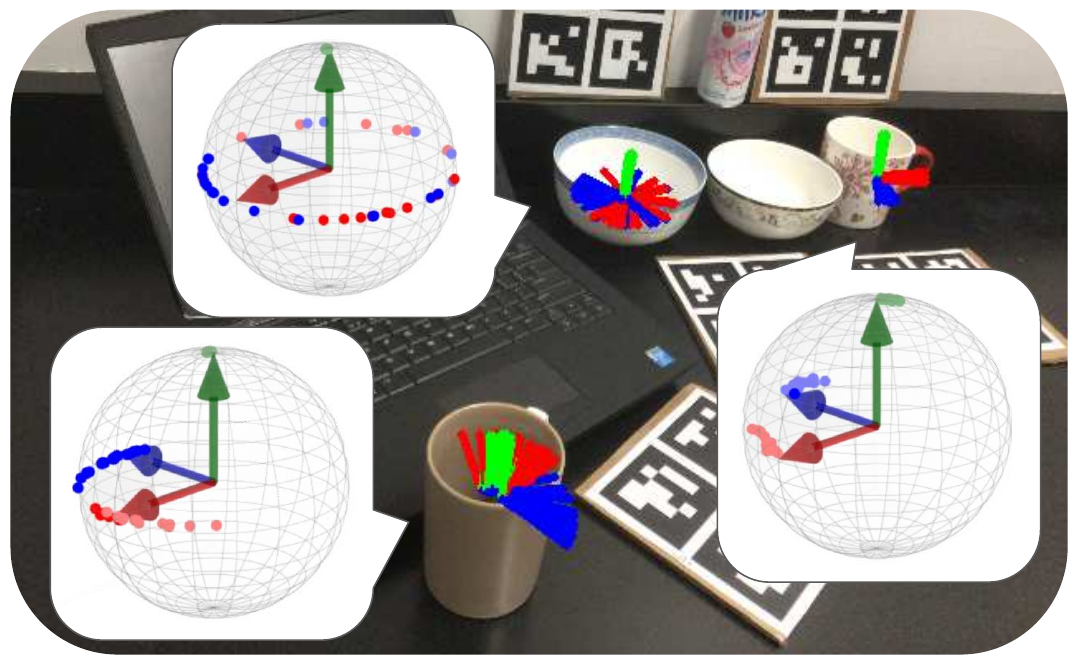}
\caption{~\textbf{Results of Multiple Estimated Poses:} A visualization of 20 estimated poses from 20 different noises. The corresponding rotations are plotted on a unit sphere to show the distribution shape of symmetry objects.}
\label{fig:symmetry}
\end{figure}

In this ablation, we qualitatively demonstrate the ability of our method to handle symmetrical objects.
As shown in \autoref{fig:symmetry}, we visualize 20 poses on the RGB image and 20 rotations on the unit sphere that are estimated from 20 different noise images.
For the bowl object, the rotations are widely distributed in one-axis since it's an axial-symmetric object. 
On the other hand, the distribution of the mug with a visible handle has well defined peaks, confirming that the handle resolves the pose ambiguity and makes it possible to establish a unique pose.
For the mug with an invisible handle, the distribution is highly correlated with the possible poses of the mug correctly identifying a possible direction of a handle (x-axis). 
These results qualitatively demonstrate the capability of our method to handle arbitrary types of symmetric objects without the need for elaborate dataset symmetry labeling and heuristic operations at training time.

\subsection{Analysis on Multimodal Input Representations}
\begin{figure}[tb]
\centering
\includegraphics[width=1.0\linewidth]{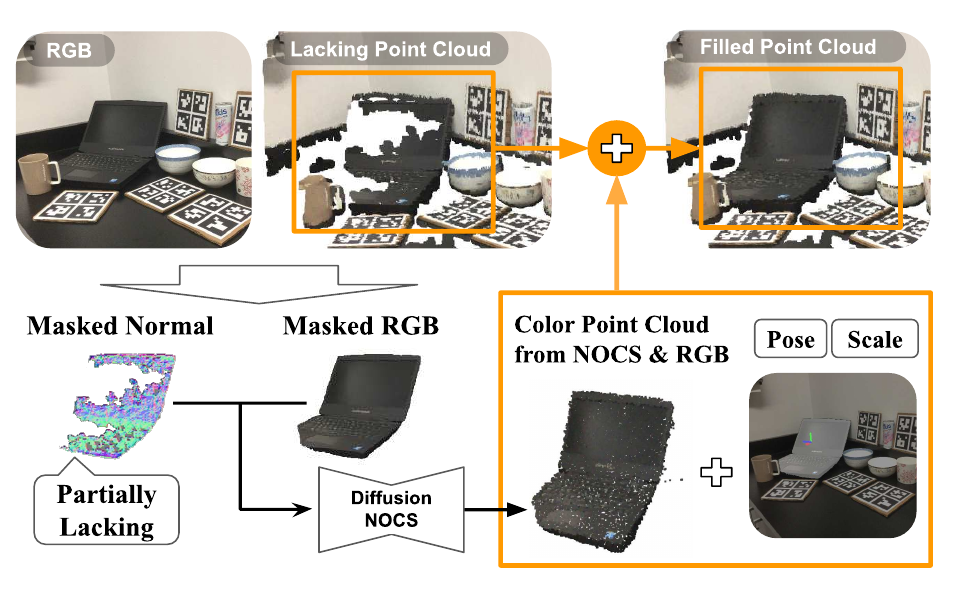}
\caption{~\textbf{Point Clouds Completion by DiffusionNOCS:} DiffusionNOCS with RGB-D information can estimate the NOCS map correctly even with a lack of depth information since RGB image can cover the information. The lacking point cloud in target objects can be reconstructed by an estimated NOCS map, pose, and scale. It's useful for various robotics applications.}
\label{fig:multimodal}
\end{figure}

It is common knowledge that structured light depth sensors often struggle to output consistent depth when it comes to reflective surfaces.
For example, the depth information of a black laptop in the NOCS Real 275 is partially lacking 
as shown in \autoref{fig:multimodal}. 
Estimating poses of such sparse point clouds is difficult or sometimes impossible, especially for methods not using RGB. 
This is also a problem for other tasks, such as detecting object-grasping locations from the point clouds.
On the other hand, our method can robustly recover the missing information thanks to its multi-modality and using both depth and RGB inputs as shown in \autoref{fig:multimodal}.
It's useful for various applications, such as not only point registration but also grasp detectors or collision avoidance.

\subsection{Ablation Study About the Parameters of PCA and Noise}

To better understand the effect of our method's parameters on the final score, we conduct an ablation study and report the results based on RGB-DINO-N inputs in \autoref{tab:ablation1}. In particular, we show how the number of (1) PCA dimensions for DINOv2 outputs and (2) noise samples affect the pose quality (mAP) and inference time (s).
Keeping the number of noise samples at 1, we observe that the performance of PCA peaks at 6 dimensions showing the best scores while maintaining minimal inference overhead. 
Changing the noise number from 1 to 6 with 6-dimensional PCA boosts the performance significantly, resulting in more than a 10 percent improvement on 10°5cm and 15°5cm metric. 
The 6 noises are chosen as a default for our pipeline from both mAP and inference time perspectives. Thus, the 5th row in \autoref{tab:ablation1} corresponds the last row in \autoref{tab:nocs_eval}.

\begin{table}[t]
    \centering
    \caption{\textbf{Ablation Results on NOCS Real 275 Benchmark With RGB-DINO-N Inputs.}}
    \resizebox{0.45\textwidth}{!}{
    \begin{tabular}{cc|rrr|c}
        \toprule
         Noise & PCA & $5 \degree 5$cm $\uparrow$ & $10 \degree 5$cm $\uparrow$ &  $15 \degree 5$cm $\uparrow$ & Time (s) $\downarrow$\\
        \hline
        1 & 3 & 27.8  & 48.7 & 56.5 & \textbf{0.372} \\ 
        1 & 6 & \textbf{30.0}  & \textbf{54.1} & \textbf{63.4} & 0.386 \\ 
        1 & 32 & 27.9  & 51.0 & 58.9 & 0.392 \\ 
        \hline
        3  & 6 &  34.5  & 65.0 & 75.2 & 0.522 \\
        6  & 6 & 35.0  & \textbf{66.6} & \textbf{77.1} & 0.855\\
        9 & 6 & \textbf{35.3}  & \textbf{66.6} & \textbf{77.1} & 1.147\\

        \bottomrule
    \end{tabular}
    }
    \label{tab:ablation1}
\end{table}

\section{Conclusion}
We propose a novel category-level dense pose estimation pipeline via diffusion models with multi-modal selectable inputs.
Thanks to its probabilistic nature, our approach is able to handle symmetrical objects without a need for special data annotations and heuristics at the training time typical for many SOTA methods. Moreover, we demonstrate that the use of multiple pose hypotheses helps to boost the performance at inference time. Outputting dense correspondences from multi-modal inputs
allow us to resolve ambiguities better and faithfully recover partial object geometry even in the presence of lacking depth.
Finally, the evaluation results demonstrate that our method significantly outperforms all SOTA baselines trained on synthetic data on the typical NOCS benchmark, and outperforms all SOTA baselines on average including the ones trained on synthetic and real data on our novel zero-shot generalization benchmark.
\section{Acknowledgement}
We would like to thank Prof. Greg Shakhnarovich, Thomas Stewart, and Dr. Vitor Guizilini for their valuable feedback.

\bibliographystyle{IEEEtran}
\bibliography{myrefs}

\vspace{12pt}

\end{document}